\pdfoutput=1

\documentclass[11pt]{article}

\usepackage{acl}

\usepackage{times}
\usepackage{latexsym}
\usepackage{graphicx}
\usepackage{booktabs}
\usepackage{mathtools}
\usepackage{amssymb}
\usepackage{amsmath}
\usepackage{amsfonts}

\usepackage[T1]{fontenc}

\usepackage[utf8]{inputenc}

\usepackage{microtype}
\usepackage{multirow}

\definecolor{kmy-color}{rgb}{0.858, 0.188, 0.478}

\definecolor{myorange}{rgb}{1.0, 0.49, 0.0}

\title{CorefDiffs: Co-referential and Differential Knowledge Flow in Document Grounded Conversations}

\author{Lin Xu$^1$, Qixian Zhou$^2$, Jinlan Fu$^1$, Min-Yen Kan$^1$, See-Kiong Ng$^1$ \\
$^1$National University of Singapore, $^2$ByteDance\\
\texttt{\{cathyxl2016,qixianzhou.mail,jinlanjonna\}@gmail.com}\\
\texttt{\{knmnyn,seekiong\}@nus.edu.sg}
}

\begin{document}
\maketitle

\begin{abstract}

Knowledge-grounded dialog systems need to incorporate smooth transitions among knowledge selected for generating responses, to ensure that  dialog flows naturally.  For document-grounded dialog systems, the inter- and intra-document  knowledge relations can be used to model such conversational flows. We develop a novel Multi-Document Co-Referential Graph (Coref-MDG) to effectively capture the inter-document  relationships based on commonsense and similarity and the intra-document co-referential structures of knowledge segments within the grounding documents. We propose CorefDiffs, a Co-referential and Differential flow management method, to linearize the static Coref-MDG into conversational sequence logic. CorefDiffs performs knowledge selection by accounting for contextual graph structures and the knowledge difference sequences. CorefDiffs significantly outperforms the state-of-the-art by 9.5\%, 7.4\% and 8.2\% on three public benchmarks. This demonstrates that the effective modeling of co-reference and knowledge difference for dialog flows are critical for transitions in document-grounded conversation\footnote{The source code has been released at https://github.com/\\cathyxl/coref-diffs}.
\end{abstract}

\section{Introduction}
\begin{figure}[t]
\centering
  \includegraphics[width=0.48\textwidth]{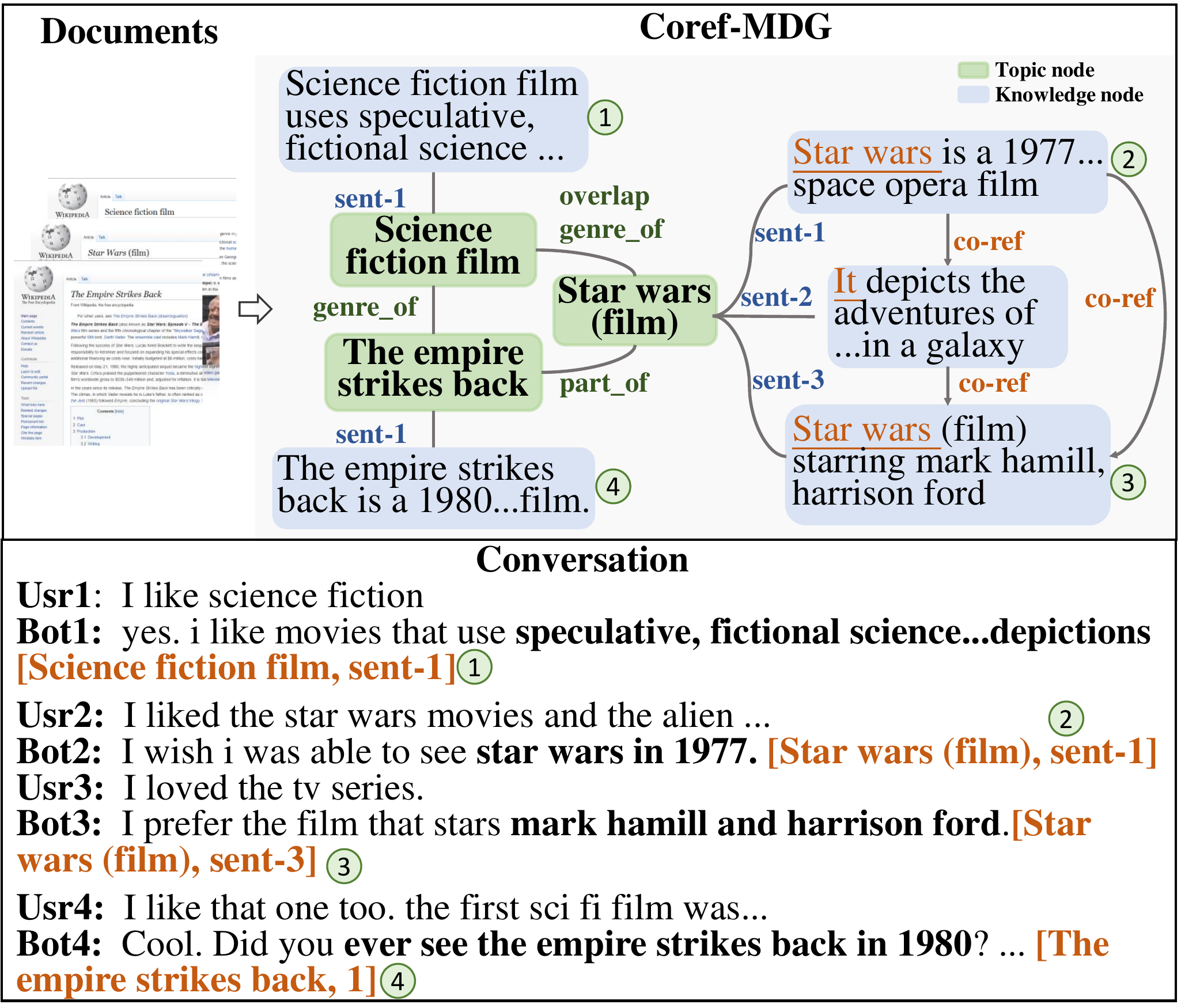}
\caption{Co-Referential Multi-Document Graph (Coref-MDG). Topic vertices correspond to documents and are connected by commonsense/word overlap relations. Knowledge vertices are connected with its topic vertex by its document sentence index, e.g. \emph{sent-1}, and connected to each other by co-reference (co-ref) relations. The Bot' s utterances are followed by its topic and knowledge segment, e.g. [Science fiction film, sent-1]. }

\label{fig:intuition}
\end{figure}
Document-grounded conversations~\cite{moghe-etal-2018-towards,dinan2018wizard,dialdoc-2021-document} is a core class of knowledge-grounded dialogs that leverage text-based knowledge segments from documents to generate informative dialog responses. This task is typically divided into two sub-tasks, given the dialog history~\cite{dinan2018wizard}: namely, knowledge selection and response generation. Knowledge selection, which determines the content of the generated responses~\cite{moghe-etal-2018-towards, dinan2018wizard}, is the crucial sub-task for dialog flow management as it leads to the manifestation of knowledge transition~\cite{meng2020dukenet}, essential for naturalistic engaging conversations.

Most existing studies on document-grounded conversations \cite{lian2019learning, zheng-etal-2020-difference,zhao-etal-2020-knowledge-grounded}  treat knowledge selection as a matching problem between the dialog context and individual knowledge segments, independently. However, for document-grounded conversations, we posit that there is an implicit alignment between the background knowledge and conversation logic which can be learned from the underlying structural relationships of the knowledge segments within and between the grounding documents.  For example,  the conversation in Figure~\ref{fig:intuition} exhibits document-level topic flow, from \textit{science fiction} -> \textit{star wars}-> \textit{the empire strikes back}, and deep dives into the specifics of the \textit{star wars} document (Turns 2 to 3).

\begin{figure}[t]
\centering
  \includegraphics[width=0.48\textwidth]{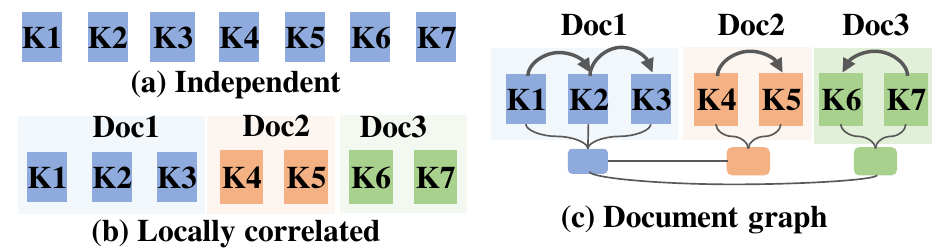}

\caption{Exploiting knowledge segment relationships. k1-7 represent knowledge segments and doc1-3 are the grounding documents they belong to.
}
\label{fig:compare}
\end{figure}

To effectively exploit the  relationships of the knowledge segments to guide  dialog flows  would require a thorough comprehension of the intra-document discourse structures and inter-document relationships for the knowledge selection process.  Existing works either ignore such relations (as illustrated in Figure~\ref{fig:compare} (a)), or    
exploit limited local correlations (as depicted in Figure~\ref{fig:compare} (b)), for example by encoding knowledge segments within passage context \citet{wu-etal-2021-dialki} . 
In this work, we propose to capture both intra- and inter-document relationships of the knowledge segments (Figure~\ref{fig:compare} (c)) in the grounding documents to guide the smooth and natural knowledge selection and transitions for document-grounded conversations.  However, how to apply such a static knowledge graph to dialog flow management has always been a problem. 
Many previous studies~\cite{moon-etal-2019-opendialkg, xu2021enhancing, xu2021coherent} have used graph structures to constrain search (e.g. confining the next topic to neighboring areas), but have also ignored deeper integration of dialog contexts and knowledge graphs, such as optimal knowledge representation to capture dialogue flow information. 

Based on the considerations above, we propose to first capture the inter- and intra-document knowledge relationships as a heterogeneous document graph, and then exploit the graph effectively for dialog flow management through fine-to-coarse contextualization --- from the local word-level knowledge attentions, to knowledge interactions in document graphs, and finally to the knowledge transition flow along dialogue turns. Specifically, we design a two-level document graph consisting of topic (i.e. document) and knowledge vertices connected by inter- and intra-document relations (Figure~\ref{fig:intuition}). The topic vertices correspond one-to-one to the grounding documents, while the knowledge vertices refer to the knowledge segments from each document.  
The knowledge vertices are connected to the corresponding topic vertices they belong to. Meanwhile, the graph connects the knowledge segments within the same document by their co-referential mentions, and the documents are connected based on similarity or commonsense relationships. Hence we call the graph Multi-Document Co-referential Graph~(Coref-MDG). 

We then propose our CorefDiffs method which leverages Coref-MDG's graph structure and integrates dialog flow for knowledge contextualization and selection.  CorefDiffs focuses on the inter-turn knowledge difference flow in the dialog histories by means of a novel differential linearization module.

Our contributions in this paper can be summarized as follows.
1) We develop Coref-MDG, a novel multi-document graph structure incorporating co-referential mentions.    
When leveraged in guiding document-grounded conversations in our CorefDiffs methodology, it empirically outperformed alternative graph structures;
2) We systematically study the different kinds of inter- and intra-document relations and show that document-level semantics, such as co-reference and sentence order, are significant factors for knowledge selection (Sec. \ref{ssec:analysis});
3) Our CorefDiffs achieves state-of-the-art on WoW, Holl-E, multidoc2dial and CMU-DOG datasets, 
for both knowledge selection and response generation tasks.

\begin{figure*}[t]
\centering
  \includegraphics[width=0.95\linewidth]{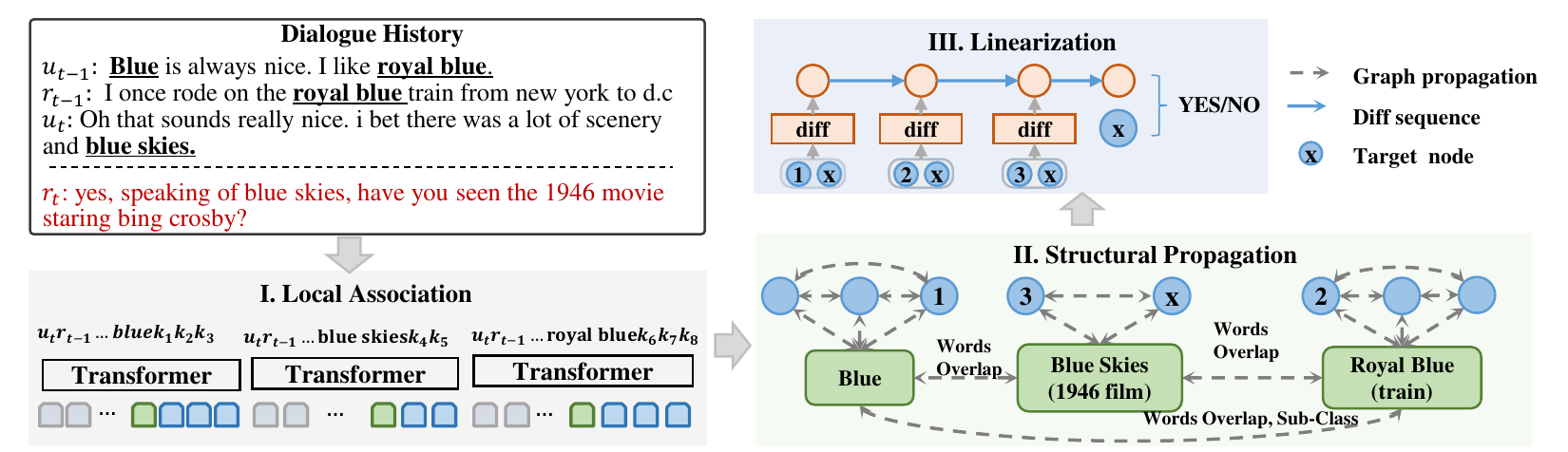}
\caption{CorefDiffs Architecture. Green round rectangles and blue circles are topic and knowledge vertices, respectively. Steps I to III contextualize knowledge in an fine-to-coarse manner: first, I) by vertex embedding by BERT; then, II)  by propagating Coref-MDG information, and finally III) by linearizing the knowledge representations, according to the dialog's historical knowledge sequence.
}
\label{fig:framework}
\end{figure*}
\section{Related Work}


\noindent\textbf{Document-grounded dialog Systems.}
Early works on document-grounded dialog systems~\cite{ghazvininejad2018knowledge} focused on generating responses directly by copying words from the external documents.  The subsequent availability of datasets with knowledge annotations~\cite{dinan2018wizard,moghe-etal-2018-towards} has led to the separation of the tasks of knowledge selection and response generation. For knowledge selection, most works~\cite{dinan2018wizard,lian2019learning, zheng-etal-2020-difference,zhao-etal-2020-knowledge-grounded,meng2021initiative} in document-grounded conversations directly modeled correlations between dialog contexts and knowledge through independent matching and optimized the correlations by modeling knowledge sequence~\cite{kim2019sequential}, increasing knowledge informativeness~\cite{zheng-etal-2020-difference} or distinguishing initiative roles~\cite{meng2021initiative}. A recent work~\cite{wu-etal-2021-dialki} boosted knowledge selection by encoding knowledge within the passage context, which demonstrates the importance of exploiting knowledge relations. Our work further explores more effective document structures and connections for this task. There is also an unpublished paper that used document semantic graphs~\cite{lienhanced}, while our work considers end-to-end integration of document graph and dialog flow which gives better result compared to theirs.

\noindent\textbf{Knowledge Graph for Conversations.}
Knowledge graphs were also often used in dialog management, such as dialog transition graphs~\cite{xu2019end, xu-etal-2020-conversational} constructed from common transitions present in a dialog corpus and off-the-shelf commonsense graphs~\cite{zhou2018commonsense}. There were also some works~\cite{liu-etal-2019-knowledge, xu2021enhancing} transforming unstructured text into structures or combining triplets and texts into graphs. For example, ~\cite{xu2021enhancing} constructed key phrases into graphs according to their order in stories. Interestingly, to the best of our knowledge, co-reference mentions have not been considered in such document graph construction  although it has been proved critical in learning language models for reasoning intensive NLP tasks~\cite{dasigi-etal-2019-quoref, ye-etal-2020-coreferential}. To apply knowledge graphs for dialog, many existing works~\cite{xu2020knowledge, xu-etal-2020-conversational} used the graph structures to confine the search space and optimized selection through hand-crafted rewards. In contrast, we incorporate the knowledge graph into dialog management by  learning knowledge representations from the graph structure.

\noindent\textbf{Sequence Learning in dialog.}
Sequence learning is essential for conversations. Several studies~\cite{kim2019sequential,zhan-etal-2021-augmenting} explored the historical knowledge sequence to select knowledge  for document-grounded dialog. For example, ~\cite{kim2019sequential} captured knowledge sequence by a latent variable, while ~\cite{zhan-etal-2021-augmenting} further proposed to learn abstract topic sequence to mitigate the issues of knowledge sparsity and knowledge transition noise. Inspired by the importance of exploiting knowledge difference~\cite{zheng-etal-2020-difference}  for informative dialog, we extend the use of dialog knowledge differences into sequences, thus capturing the knowledge shift patterns from turns with longer distances as well as the sequential patterns of knowledge transitions in a dialogue.

\section{Approach}

Figure~\ref{fig:framework} shows the overall architecture of our approach. As shown in the Dialog History part, in each data sample, given a dialog history $U=\{u_{t-l}, r_{t-l},...,r_{t-1}, u_{t}\}$ of $l$ turns and a set of grounding documents $\mathcal{D}=\{d_1,.., d_i, ..., d_{|\mathcal{D}|}\}$, where $u_*$ and $r_*$ are utterances from the user and chatbot, respectively. $d_i=\{k^i_1, k^i_2, ..., k^i_{|d_i|}\}$ is a document containing a bunch of knowledge segments, our task is to select the most appropriate knowledge segment from the grounding documents $\mathcal{D}$ (i.e. the knowledge selection subtask) and generate the chatbot's next response $r_t$ based on the selected knowledge (i.e. the response generation subtask). Each grounding document $d_i$ has a phrase $t_i$ as its topic. For example, the document of wikipage blue has the topic phrase \textit{blue}.

\subsection{Coref-MDG Construction}
\label{ssec:graph_construction}
We devise a Multi-Document Co-referential Graph (Coref-MDG) to capture the inter-document and the intra-document relations. Each data sample gets a specific Coref-MDG, denoted as $\mathcal{G}=\{\mathcal{V},\mathcal{E}\}$, where $\mathcal{V}$, $\mathcal{E}$ are vertices and edges respectively.

\noindent\textbf{Vertices $\mathcal{V}$.} 
Our Coref-MDG consists of two types of Vertices: topic and knowledge vertices, as shown in Figure~\ref{fig:intuition}. Each topic vertex represents one of document $d_i$ from $\mathcal{D}$ while each knowledge vertex represents a knowledge segment $k^i_j$ from a document $d_i$, hence in total $M = |\mathcal{D}|$ topic vertices and $N=|d_1|+...+|d_{|\mathcal{D}|}|$ knowledge vertices. 

\noindent\textbf{Edges $\mathcal{E}$.} There are also multiple types of edges in Coref-MDG. We can generally divide these edges into three categories according to their vertices: \textbf{1) edges between topics and knowledge vertices}; \textbf{2) topic edges} --these are inter-document or inter-topic edges between topic vertices;  \textbf{3) knowledge edges} --intra-topic edges amongst knowledge vertices.  For the first category, we simply use the order index of segment $k^i_j$ appearing in its corresponding document $d_i$ as the edge type, denoted as \texttt{sent\_j} edge, thus knowledge vertices under different topics are not connected in Coref-MDG. The remaining two types of edges are constructed as follows.

\subsubsection{Topic Edges}
We posit that topic transitions in human-to-human conversations are likely to be based on the similarity or commonsense relations between two topics, such as from \emph{sci-fi movie} to \emph{sci-fi novel} (similarity), or from \emph{UK} to \emph{London} (commonsense). We introduce two corresponding types of topic edges for such topic transitions.

\noindent\textbf{Word Overlap.}
We use the word overlaps between two topics (or documents) to measure their similarity. Specifically, we obtain the lemmas of topic phrases by spaCy\footnote{https://spacy.io/, MIT License} and judge whether the two topics have at least one identical lemma so as to determine whether these two topics vertices have a \texttt{word\_overlap} edge.

\noindent\textbf{Commonsense.}
Since the knowledge backend of the WoW~\cite{dinan2019wizard} came from the Wikipedia corpus, we use the WikiData\footnote{https://www.wikidata.org/wiki/Wikidata:Main\_Page} to obtain commonsense relations between topics. We only collected relations for the topics in the training set and for cimplicity, we kept the high-frequency relation types, for example \texttt{city\_of}, while uniformly treating the low-frequency relation types as \texttt{others}.

\subsubsection{Knowledge Edges}
For the intra-document knowledge relations, we introduce the \texttt{coreference\_link} edge. For each topic (i.e. document), the co-reference links (referring paths) within the corresponding document $d_i$ can be extracted by a co-reference resolution model.~\footnote{https://github.com/huggingface/neuralcoref, MIT License} For each co-reference link, every knowledge segment on this link is connected to its mentions by a \texttt{coreference\_link} edge. Aside from our proposed co-reference edges, we also model two other knowledge edge type for comparison, \texttt{common\_entity} and \texttt{partial\_order}. 
The former connects knowledge segments that share entities, while the latter captures knowledge segment's partial order. We will show later that co-reference performs best for dialog flow management. 

\subsection{Structural Propagation and Linearization}
Next, we introduce how we contextualize each vertex in a dialog's Coref-MDG with both the graph structure and dialog flow.
\subsubsection{Node Initialization}
Following ~\cite{karpukhin-etal-2020-dense, cheng-etal-2020-probabilistic, wu-etal-2021-dialki}, we adopt BERT~\cite{devlin-etal-2019-bert} to obtain the text representations to initialize topic and knowledge vertices, as shown by the Step~\textbf{I} in Figure~\ref{fig:framework}. Specifically, we concatenate the dialog context $U$ with each grounding document's topic phrase and knowledge segments, and feed them into the BERT encoder to get their associated representations. The concatenated input for a document $d_i$ is thus:
\begin{equation}
    [\mathrm{cls}] \hat{U}_t[\mathrm{sep}]t_i[\mathrm{cls}]k^i_1 ... [\mathrm{cls}]k^i_{|d_i|}[\mathrm{sep}]
\end{equation}
\noindent where $\hat{U}=[\mathrm{usr}]u_{t}[\mathrm{agt}]r_{t-1} ... [\mathrm{usr}]u_{t-l}$ is the spliced dialog context, and the role symbols $[\mathrm{usr}]$ and $[\mathrm{agt}]$ indicate utterances from the user or agent turn. We use the hidden state of the first $[\mathrm{cls}]$ token $\mathbf{t_i}$ (note that we use \textbf{bold} here to refer to the representation of, in this case, topic phrase  $t_i$) as the initialized representation for the topic vertex of $d_i$. Similarly, the outputs of the subsequent $[\mathrm{cls}]$ tokens, denoted as $\{\mathbf{k^i_1}, \mathbf{k^i_2}, ..., \mathbf{k^i_{|d_i|}}\}$, are gathered and used to initialize the corresponding knowledge vertices of $d_i$. The process is formulated as:
\begin{equation}
    \mathbf{t_i}, \mathbf{K^i} = \text{BERT}(U_t, d_i),i\in[1,{|\mathcal{D}|}]
\end{equation}
where $\mathbf{t_i},\mathbf{k^i_j} \in \mathbb{R}^{d_{init}}, \mathbf{K^i}= \{\mathbf{k^i_j}\}_{j=1}^{|d_i|}$. In this way, we obtain the initialized vertex embedding for a Coref-MDG as $\mathbf{H^0}=\{\mathbf{t_i};\mathbf{K^i}\}_{i=1}^{|\mathcal{D}|} \in \mathbb{R}^{(M+N) \times d_{init}}$.

\subsubsection{Residual Graph Propagation}
Knowledge transitions in document-grounded dialogs can be divided into two types namely, transition across different documents~(\textbf{out-topic}) and  within the same document~(\textbf{intra-topic}). Transitions across different documents occur between topic vertices in our Coref-MDG and usually requires multi-hop reasoning. We use the residual graph propagation~(Step~II in Figure~\ref{fig:framework}) to model such transitions in Coref-MDG. Specifically, we devise a variant of Relational Graph Attention Layer(RGAT)~\cite{busbridge2019relational} layer with concatenated residual connection~\cite{he2016deep}, named \textsc{Res-RGAT}. This layer facilitates the deeper multi-hop information propagation by avoiding information loss and the over-smooth problem~\cite{oono2019graph}. The output $\mathbf{H^{out}} \in \mathbb{R}^{|\mathcal{G}| \times d_{out}}$ of one \textsc{Res-RGAT} is the concatenation of the propagated results and the input $\mathbf{H^{in}} \in \mathbb{R}^{|\mathcal{G}| \times d_{in}}$, which is formulated as:
\begin{equation}
\mathbf{H^{out}} = W[\mathbf{H^{in}}, \textsc{RGAT}(\mathbf{H^{in}}, \mathbf{R}, \mathcal{E}, \mathcal{G})]\\
\end{equation}
\noindent where $\mathbf{R} \in \mathbb{R}^{E \times d_e}$ is the embedding look-up table for all the edge types in $\mathcal{E}$, $E$ is the number of edge types, and $W\in \mathbb{R}^{d_{out} \times 2d_{in}} $ is used for dimension transform. We stack $n$ layers of \textsc{Res-RGAT} to do enough propagation based on empricially determined $n$. With $\mathbf{H^0}$ as input, we obtain the propagated outputs for all vertices as $\mathbf{H^G} \in \mathbb{R}^{(M+N)\times d_G}$.

\subsubsection{Differential Linearization}
To integrate the dialog flow information into the knowledge representations after graph propagation, we propose a novel Differential Linearization ~(Step~III in Figure~\ref{fig:framework}) method. While knowledge sequence has been used for knowledge selection in dialog~\cite{kim2019sequential}, knowledge shift sequence (or shifting sequence), defined as the sequence of knowledge differences within each consecutive turns, is a relatively novel notion for this task. We argue that the shifting sequence is a more useful feature for learning and predicting knowledge transitions since it focuses on the difference and interaction between knowledge, leading to sharper features. It also captures the transition patterns from turns using varying distances to the current turn to further aid in the  selection.

To construct the shifting sequence, we first obtain the knowledge/topic vertices that appeared in the previous chatbot turns (since we note that the labels of the user turns are inaccessible in practice). By collecting these knowledge/topic vertices' representations from $\mathbf{H}^G$, we can get the sequence $S=\{\mathbf{h}^G_{t-\tau}, ..., \mathbf{h}^G_{t-1}\}$ for knowledge and topic vertices, respectively. Here $\tau$ is the length of turns. 
Since we  treat topic and knowledge vertex sequence identically, we will refer to them as simply vertices in the following discussion. We compare the vertex $i$ with these historical vertices in $S$ with a comparison function $\mathcal{F}$ to get the differential sequence for vertex $i$. By doing this for all vertices, we get $M+N$ such sequences: 
\begin{equation}
    \{\mathcal{F}(\mathbf{h}^G_{t-\tau},\mathbf{h}^G_i), ...,\mathcal{F}(\mathbf{h}^G_{t-1},\mathbf{h}^G_i)\}^{M+N}_{i=1}
\end{equation}
$\mathcal{F}$ computes the interaction between two vectors $\mathbf{a},\mathbf{b} \in \mathbb{R}^d$ by element-wise difference and  product, defined as $\mathcal{F}(\mathbf{a},\mathbf{b}) = [\mathbf{a}-\mathbf{b};\mathbf{a}\odot \mathbf{b}]$~\cite{chen-etal-2017-enhanced}.

With the sequence for vertex with index $i$, we finalize its representations in sequential transition dependency. Specifically, we feed each sequence into a stacked GRU~\cite{cho2014learning} cells and use the last hidden state as the final linearized vertex representation. We concatenate the graph representation of vertex $h^G_i$ and the linearized output: 
\begin{equation}
    \mathbf{h}^D_i=[\text{GRU}(...,\mathcal{F}(\mathbf{h}^G_{t-1},\mathbf{h}^G_i));\mathbf{h}^G_i] \in \mathbb{R}^{2d_G}
\end{equation}
The vertex representation in graph after linearization is $\mathbf{H^D} \in \mathbb{R}^{(M+N)\times 2d_G}$, which will then be used to predict the next topic and knowledge segment.

\subsection{Training}
Note that topic selection is an auxiliary task in our framework, apart from the knowledge selection. As such, we split the representations for topic vertices and knowledge vertices from $\mathbf{H}^D$ and obtain $\mathbf{H}^D_{tpc}\in \mathbb{R}^{M\times2d_G}$ and $\mathbf{H}^D_{knl}\in\mathbb{R}^{N\times2d_G}$, respectively. $\mathbf{H}^D_{tpc}$ is fed into a linear layer to obtain the topic selection scores. For the knowledge vertices, we further include their connected topic vertex representations and the in-between edge embedding to calculate the knowledge selection scores similarly with a linear layer.

Following \citet{wu-etal-2021-dialki}, we implement the history loss as an auxiliary objective function in our framework to further utilize the dialog history information. Finally, the overall objective function we adopt is formulated as follows:
\begin{equation}
\begin{aligned}
    \mathcal{L}&=\mathcal{L}_{knl} + \mathcal{L}_{tpc} + \mathcal{L}_{hist}\\
    \mathcal{L}_{hist}&=\frac{1}{2l}\sum_{hi=1}^{l}(\mathcal{L}^{hi}_{knl} + \mathcal{L}^{hi}_{tpc})
\end{aligned}
\end{equation}
\noindent where $l$ is a hyperparameter representing the history length, $\mathcal{L}_{knl}$ and $\mathcal{L}_{tpc}$ are knowledge and topic losses, respectively. All of the classification objective functions in $\mathcal{L}$ are standard cross-entropy.

\section{Experiments}
\label{sec:experiments}
\noindent\textbf{Datasets. }
We validate our method on four public benchmarks for document-grounded conversation, WoW~\cite{dinan2018wizard}, Holl-E~\cite{moghe-etal-2018-towards}, CMU-DoG~\cite{zhou-etal-2018-dataset} and MultiDoc2Dial~\cite{feng-etal-2021-multidoc2dial}. The dataset statistics are summarized in Table~\ref{tab:dataset}. We first conduct knowledge selection with our Coref-Diffs method and then feed the selections and dialogue history into text generation models to compare the final responses. 

\begin{table}[t]
\centering
\renewcommand\tabcolsep{2.2pt}
\resizebox{\linewidth}{!}{
\begin{tabular}{lcccc}
\toprule
\textbf{Method} & \textbf{ \#dialogs}& \textbf{ \#Avg Turns} & \textbf{Domain} & \textbf{Document} \\ 
\midrule      
WoW & 22311  & 9 & Open Domain & Multiple \\
Holl-E   & 9071 &  10 &  Movie & Single \\
CMU-DoG & 4112  &  31  &  Movie & Single\\
MultiDoc2Dial  &  4796 &  14  & Info Seek  &  Multiple  \\   
\bottomrule
\end{tabular}}
\caption{Dataset statistics.}
\label{tab:dataset}
\end{table}



\noindent\textbf{Evaluation metrics.}
We focused on evaluating the knowledge selection sub-task for the document-grounded dialog system, based on the knowledge and topic selection accuracies, denoted as \textbf{KL} and \textbf{TP}, respectively. We also explore the knowledge selection accuracy of all intra-topic data samples, whose knowledge transitions are within the same topic, denoted as \textbf{In-TP}. As for evaluating the  sub-task of response generation given the dialog context and selected knowledge, we calculate the overlap of the generated response and the ground-truth with the unigram-F1(\textbf{uF1}) and bigram-F1(\textbf{bF1}).

\noindent\textbf{Baselines.}
For the two commonly used datasets, WoW and Holl-E, we split the baselines into three categories based on their text encoder types. (i) \textbf{Non-Pretrained encoder}: Transformer+MemNet~\cite{dinan2018wizard} is the baseline released with the dataset WoW. DiffKS(RNN)~\cite{zheng-etal-2020-difference} incorporates the knowledge difference feature in knowledge selection. (ii) \textbf{BERT encoder}: BERT+PoKS, a variant of PoKS with BERT~\cite{devlin-etal-2019-bert} encoder, learns knowledge selection by posterior knowledge distribution. SLKS~\cite{kim2019sequential} captures historical knowledge sequence with a latent variable. PIPM~\cite{chen-etal-2020-bridging} improves SLKS by addressing the problem of missing posterior distribution in test phase. CoLV~\cite{zhan-etal-2021-colv} includes two collaborative variables for knowledge selection and response generation. KnowledGPT~\cite{zhao-etal-2020-knowledge-grounded} optimizes knowledge grounded dialog task by the pre-trained BERT encode and GPT-2~\cite{radford2019language}. (iii) \textbf{Passage-level BERT encoder}: DIALKI~\cite{wu-etal-2021-dialki} encodes knowledge at passage level to capture knowledge segment relations as we do in CorefDiffs. For response generation, given that the above-mentioned methods adopted different generators,  we uniformly replaced their generators with a prompt-based generator PrefixTuning~\cite{li-liang-2021-prefix} for a fair comparison, thus forming the baselines with "*" in Table~\ref{tab:res_generation}. For MultiDoc2Dial and CMU-DoG, we compare our method with the current state-of-the-art DPR+RAG~\cite{lewis2020retrieval} and DoHA~\cite{prabhumoye-etal-2021-focused}, respectively. The  generators used are fine-tuned BART-large.

\subsection{Implementation Details.}
The BERT-base models in all our experiments used the Huggingface Transformers\footnote{https://github.com/huggingface/transformers}~\cite{wolf-etal-2020-transformers}. We trained the model with Adam~\cite{kingma2015adam} optimizer with initial learning rate 1e-5. A linear scheduler with a warm-up strategy in 5k steps was used. The maximum history length $l$ was empirically set to 4 for WoW, 2 for Holl-E, 3 for CMU-DoG and 4 for MultiDoc2Dial to achieve the best performance. The number of the stacked Res-RGAT was set to 2. It took around 5 and 10 epochs to achieve the reported performance by 4 nvidia V100 GPUs. We will release all the codes and the hyper-parameters settings for reproduction.

\subsection{Automatic Evaluations}
\label{sec:ablation}
\begin{figure*}[t]
\centering
  \includegraphics[width=0.98\linewidth]{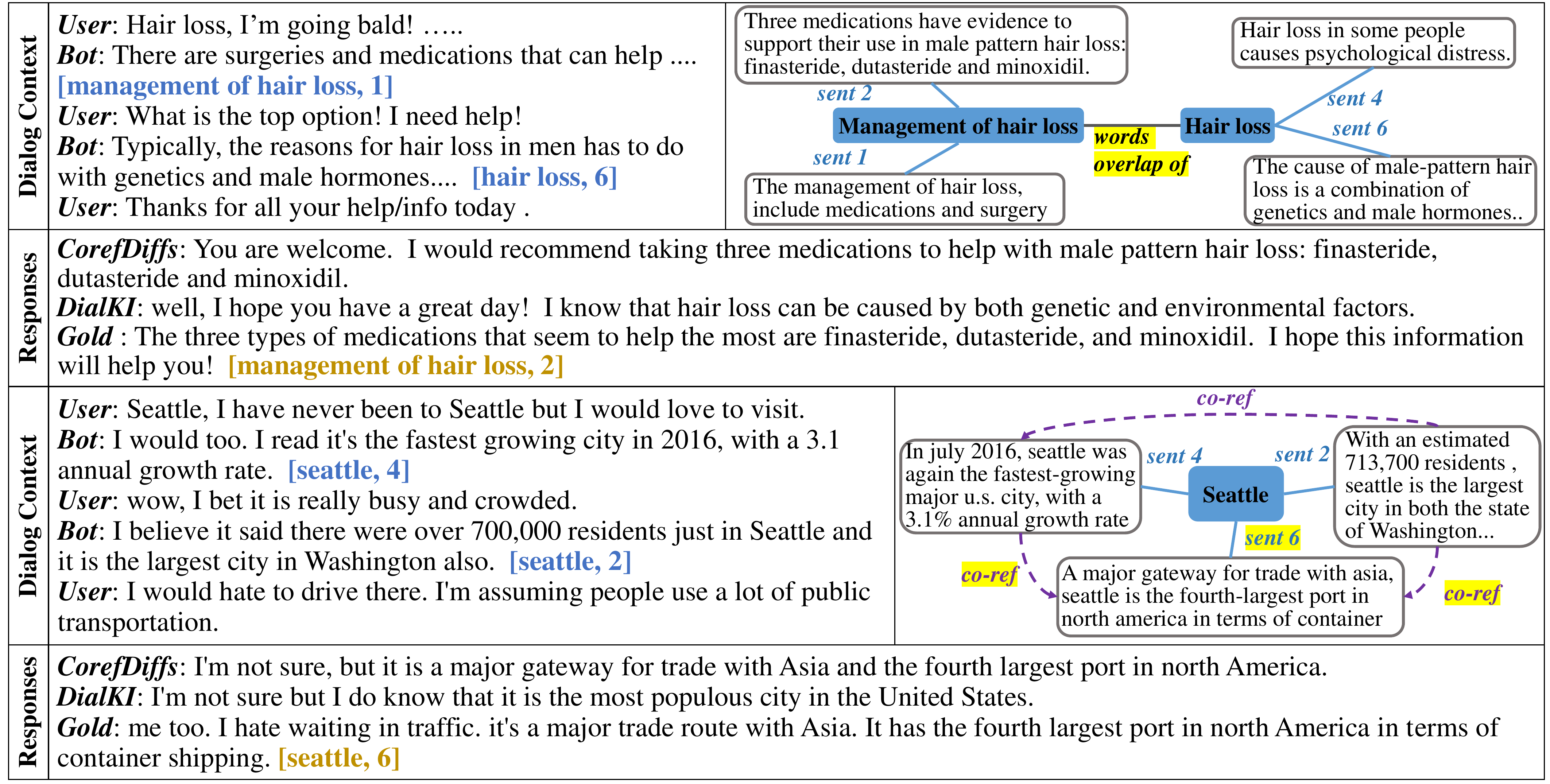}
\caption{Two generation examples from WoW. The bold words in "[]" indicate the knowledge. For example, \textbf{[hair loss, 6]} represents the 6-th knowledge sentence in the document with topic \textbf{hair loss}. Our method chose the right knowledge for both examples compare to DIALKI owing to the well-designed graph structure.}
\label{fig:visual}
\end{figure*}
\noindent\textbf{Knowledge Selection.}
\begin{table}[t]
\centering
\renewcommand\tabcolsep{3.5pt}
\resizebox{0.85\linewidth}{!}{
\begin{tabular}{lcccccc}
\toprule
\multirow{1}{*}{\textbf{Method}} & \textbf{WoW (Seen)}& \textbf{WoW (Unseen)} & \textbf{Holl-E} \\ 
\midrule
TMN              &  22.5  & 12.2   &   22.7   \\
DiffKS(RNN)      &  25.6  &  18.6   &   33.5 \\
BERT+PoKS  &  25.5  &  14.1   &   27.6     \\
SLKS       &  26.8  &  18.3   &   29.2   \\
PIPM             &  27.8  &  19.7   &   30.7     \\
CoLV             &  30.1  &  18.9   &   32.7   \\
DukeNet          &  26.4  &  19.6   &   30.0    \\
KnowledGPT       &  28.0  &  25.4   &   -    \\
\midrule
DIALKI           &  32.9  &  35.5   &   -   \\
CorefDiffs       &  \textbf{42.4}  &  \textbf{41.4}   &   \textbf{40.9} \\ 
 \quad w/o Diff-Seq   &  40.8  &  39.5   &  39.7   \\ 
 \quad w/o Diff       &  40.9  &  40.1   &  40.1   \\ 
 \quad w/o Res-RGAT   &  35.5  &  36.5   &  39.5   \\ 

\bottomrule
\end{tabular}
}
\caption{
The knowledge selection results measured by accuracy on WoW and Holl-E. 
}
\label{tab:know_selection}
\end{table}
The knowledge selection results on the four datasets are presented in Table~\ref{tab:know_selection} and ~\ref{tab:cmudog_multidoc2dial_results}. CorefDiffs significantly outperforms all other methods regardless of the encoders they used. Compared to the best performance achieved by DIALKI, CorefDiffs improves by $9.5\%$ and $5.9\%$ and is the first to achieve knowledge accuracy over 40\% on both the WoW Test Seen and Unseen sets. For Holl-E, CorefDiffs also performs the best, with gains of at least $7.4\%$ in knowledge selection. For MultiDoc2Dial our method outperforms state-of-the-art by 8.2\% in knowledge selection accuracy. CMU-DoG has no ground-truth knowledge, so we only report generation results.
The substantial enhancements across all datasets strongly suggest that CorefDiffs has benefited from modeling document structures and knowledge relations in the grounding documents with differential dialog flow learning.

\noindent\textbf{Response Generation.}
\begin{table}[t]
 \renewcommand\tabcolsep{3pt}
\centering
\resizebox{0.9\linewidth}{!}{
\begin{tabular}{lcccccc}
\toprule
\multirow{2}{*}{\textbf{Method}} & \multicolumn{2}{c}{\textbf{WoW (Seen)}}& \multicolumn{2}{c}{\textbf{WoW (Unseen)}} & \multicolumn{2}{c}{\textbf{Holl-E}}\\ 
\cmidrule(lr){2-3}\cmidrule(lr){4-5}\cmidrule(lr){6-7}
                         & \textbf{uF1} & \textbf{bF1}  & \textbf{uF1} & \textbf{bF1} & \textbf{uF1} & \textbf{bF1} 
                         \\ 
\midrule
SLKS(TM+Copy)         & 19.3 & 6.8  & 16.1 & 4.2  &  29.2 &  22.3   \\
DukeNet(TM+Copy)      & 19.3 & 6.3   & 17.1 & 4.7  &  30.6 & 23.1 \\
SLKS*                  & 20.2 & 7.3   & 17.5 & 5.3   &   -  &  -  \\
DiffKS*        & 21.5 & 7.6   & 20.0 & 6.3   &  30.7 & 23.9    \\
KnowledGPT*    & 22.0 & 8.2   & 20.8 & 7.4   &  -  &  -   \\
DIALKI*        & 22.0 & 8.0  & 22.2 & 8.1   &  -   & -   \\
\midrule
CorefDiffs        & \textbf{25.2} & \textbf{10.7}   & \textbf{25.8}  & \textbf{10.8}  &  \textbf{38.4} & \textbf{31.8} \\ 
\bottomrule
\end{tabular}}
\caption{Response generation results on WoW and Holl-E. Methods with `(TM+Copy)' and `*' used generator Transformer + Copy mechanism and PrefixTuning.`-' indicates the method didn't do experiment on the dataset.}

\label{tab:res_generation}
\end{table}
Tables~\ref{tab:res_generation} and ~\ref{tab:cmudog_multidoc2dial_results} show the results of response generation on all the four datasets. We applied PrefixTuning~\cite{li-liang-2021-prefix} to generate responses with the corresponding dialog context and selected knowledge as the input for WoW and Holl-E, while for MultiDoc2Dial and CMU-DoG, we followed previous works using BART-Large. The PrefixTuning obtained a comparable performance with fewer learnable parameters and extrapolated better to unseen topics than fine-tuning method. Again, CorefDiffs obtains best performance in all generation metrics on four datasets, which we attribute to the large margins on knowledge selection.

\noindent\textbf{Ablation Study.}
To study the impact of each of the modules in CorefDiffs, we conduct three experiments, as shown in the lower part of Table~\ref{tab:know_selection}. For \emph{w/o Diff-Seq}, we remove the Differential Linearization. \emph{w/o Diff} uses the normal knowledge sequence instead of the shifting sequence. \emph{w/o \textsc{Res-RGAT}} removes the Residual Graph Propagation. After removing \textsc{Res-RGAT}, we observe a steep drop in knowledge selection accuracy, which proves that knowledge representations updated by graph propagation on Coref-MDG is well aligned with knowledge distribution in next turn. This also shows that purely relying on local correlations within passage context (Step I in Figure~\ref{fig:framework}) is not as good as using higher-level document structures. In addition, \emph{w/o diff-seq} also presents lower knowledge selection accuracy, showing its importance to enhance dialog flow modeling upon graph propagation. More importantly, by comparing \emph{w/o diff-seq} and \emph{w/o Diff}, we notice that using shifting sequences outperforms normal ones, thus validating our earlier argument that shifting features are sharper and more effective for dialog knowledge flow.     

\begin{table}[t]
\centering
\resizebox{0.7\linewidth}{!}{
\begin{tabular}{lccc}
\toprule
\multirow{2}{*}{\textbf{Method}} & \multicolumn{2}{c}{\textbf{MultiDoc2Dial}}& \textbf{CMU-DOG}\\ 
\cmidrule(lr){2-3}\cmidrule(lr){4-4}
& \textbf{uF1}& \textbf{KL} & \textbf{uF1}\\
\midrule      
DoHA   & -   &-  & 22.8  \\
DPR+RAG & 33.7   & 24.9   &-  \\
CorefDiffs  &  \textbf{39.3} & \textbf{33.1}  &  \textbf{23.9}  \\   
\bottomrule
\end{tabular}
}
\caption{Experimental Results on MultiDoc2Dial and CMU-DOG.}
\label{tab:cmudog_multidoc2dial_results}
\end{table}

\subsection{Case Study}
Why does Coref-MDG work on knowledge selection?
To answer this question, we visualize two typical examples, shown in Figure~\ref{fig:visual}.
The \emph{dialog Context} rows are dialog histories, and the generated responses of different methods are listed in the \emph{Response} row. We compare our CorefDiffs with DIALKI and the Gold (ground-truth) response. The first example performed topic change from ``hair loss'' to ``management of hair loss''. CorefDiffs chose the right knowledge topic, ``management of hair loss'', while DIALKI repeated the knowledge mentioned in the earlier conversation turn. The reason is that CorefDiffs was able to do so is because it had referred to the \texttt{word\_overlap} connection between ``hair loss'' and ``management of hair loss'', whereas DIALKI did not consider such inter-topic relations. For the second example, the knowledge transition is intra-topic (knowledge in consecutive turns belonging to the same topic/document). Our method successfully predicts the right knowledge due to the co-reference relation between these knowledge sentences within the  ``seattle'' document, whereas  the response generated by DIALKI --- even with passage-level knowledge correlations encoded --- missed the long dependency from the second to the sixth sentence.
\subsection{Graph Analysis}
\label{ssec:analysis}

To study effects of the different type of relations in Coref-MDG on topic/knowledge selection accuracies. We did more experiments on WoW.
We craft 3 Coref-MDG variants lie in three categories for relations between topics. (1)~\emph{w/o TP}: removing all topic edges; (2)~\emph{w/o TP overlap}: removing the word overlap edges; (3)~\emph{w/o TP wikigraph}: removing the commonsense edges. Another three variants for exploring the relations between knowledge vertices are as follows: (4)~\emph{w/o KG}: removing all knowledge edges (that is co-reference link); (5)~\emph{+KG common entity}: applying entity edges between knowledge instead; (6)~\emph{+KG partial order}: employing partial order edges between knowledge. We also remove the sentence order edges between topic and knowledge vertices and formed a variant (7)~\emph{w/o TP-KG}. The results of the above experiments are listed in Table~\ref{tab:graph_comparison}, from which we get the following \textbf{conclusions}:

\paragraph{(i)~Coref-MDG performs the best in knowledge selection compared to other graph structures.} Removing or replacing edge types in Coref-MDG, such as the edges between topic vertices (Exp.~1-3), knowledge vertices (Exp.~4-6), or topic and knowledge vertices (Exp.~7), can cause a drop in topic or knowledge selection both on Seen or Unseen settings. Moreover, sometimes using other kind of edge leads to worse results than their absence. For example, in Exp.~4 and 5, \textit{w/o KG} performs better than \textit{+ KG common entity} in Unseen.

\paragraph{(ii)~Topic and knowledge selection accuracies are affected by their relevant relations in Coref-MDG.} In Exp.~1 and 7, without topic edges or topic-knowledge edges, the model achieves lowest TP. In Exp.~4-7, model achieves lower KL without suitable knowledge relations.  

\paragraph{(iii)~Topic and Knowledge relations also facilitate each other.} In Exp.~1 and 4 even removing topic relation or knowledge relations, the model still achieves better TP and KL compared to DIALKI~(no graph relation used).

\paragraph{(iv)~Knowledge relations improve intra-topic knowledge selection.} As shown by results in 4th and 7th columns, by comparing In-TP results in Exp.~1-3 and Exp.~4-6, after removing knowledge edges, the In-TP drops a lot, thus we conclude that relations between knowledge enhance the intra-topic knowledge selection.





\begin{table}[t]
  \renewcommand\tabcolsep{2.2pt}
\centering 
\resizebox{0.98\linewidth}{!}{
\begin{tabular}{lcccccc}
\toprule
\multirow{2}{*}{\textbf{Method}} & \multicolumn{3}{c}{\textbf{WoW Seen}}& \multicolumn{3}{c}{\textbf{WoW Unseen}}\\ 
\cmidrule(lr){2-4}\cmidrule(lr){5-7}
                        & \textbf{KL} & \textbf{TP} & \textbf{In-TP} & \textbf{KL} & \textbf{TP} & \textbf{In-TP}\ \\ 
\midrule
DIALKI       &    32.9     &  70.0   & 42.3  &    35.5       &   71.6 & 43.5 \\
CorefDiffs       &    42.4     &  76.1   & 51.1 &    41.4       &   77.7  & 49.2\\
\midrule

1.\quad w/o TP              & 42.1     & 74.0    & 50.6   &  39.8    &  75.2   & 47.2 \\
2.\quad w/o TP overlap      & 42.4     & 75.9    & 51.2   &  40.9    &  77.7   & 48.1 \\
3.\quad w/o TP wikigraph    & 42.3     & 75.9   & 50.9    &  41.1    &  77.5   & 48.8 \\
\midrule
4.\quad w/o KG              &  35.4  & 75.7  & 44.6  & 37.1  & 77.2  & 46.1 \\
5.\quad \quad + KG common entity         &  35.4  & 74.6  & 44.4  & 36.4  & 75.9  & 43.8 \\
6.\quad \quad + KG partial order  &  36.6  & 75.9  & 45.7  & 37.1  & 76.8  & 45.5 \\ 
\midrule
7.\quad w/o TP-KG           &  40.5  & 73.5  & 49.5  & 38.3  & 74.5  & 45.7\\

\bottomrule
\end{tabular}
}
\caption{Graph Comparisons in selection accuracy. 
}
\label{tab:graph_comparison}
\end{table}


\section{Conclusion}

We show the significance of utilizing the document's semantic structures and relations for managing dialog flow. We embody these relations in  our novel multi-document graph Coref-MDG which models co-referential knowledge mention links and inter-document relations. Our analysis of Coref-MDG yields insights of how the difference among intra- or inter-document relations affect the final topic and knowledge selection accuracy. For example, we find that coreference links and topic-knowledge sentence order relations are critical relations. 
We then build dynamically-sensitive dialog flows via our CorefDiffs method, which integrates the modeling of dialog difference flow with the prior knowledge represented in Coref-MDG.
CorefDiffs demonstrates that it is possible to seamlessly integrate static graph structures with dynamic dialog-specific flows, improving document-grounded conversations. 



\section*{Ethical Impact}

Document-grounded dialog technology has broad application prospects in open-domain dialog, emotional escort robots, intelligent assistants, etc. This work focuses on knowledge selection  which plays a significant role in dialog management of multi-turn dialog for document-grounded conversations.  All datasets we used in this work were privacy filtered and content moderated by the dataset authors~\cite{dinan2019wizard,moghe-etal-2018-towards}. However, advanced dialog knowledge selection techniques may also enable bots to select harmful content on the Internet and generate inappropriate or biased responses to users.  Future work should take this into consideration. 

\section*{Acknowledgements}
We thank all reviewers for their valuable comments. This work is supported by the National Research Foundation, Singapore under its AI Singapore Programme (AISG Award No: AISG-GC-2019-001-2A) and Industry Alignment Fund – Pre-positioning (IAF-PP) Funding Initiative. Any opinions, findings and conclusions or recommendations expressed in this material are those of the author(s) and do not reflect the views of the National Research Foundation, Singapore.


\begin{thebibliography}{43}
\expandafter\ifx\csname natexlab\endcsname\relax\def\natexlab#1{#1}\fi

\bibitem[{Busbridge et~al.(2019)Busbridge, Sherburn, Cavallo, and
  Hammerla}]{busbridge2019relational}
Dan Busbridge, Dane Sherburn, Pietro Cavallo, and Nils~Y Hammerla. 2019.
\newblock Relational graph attention networks.
\newblock \emph{arXiv preprint arXiv:1904.05811}.

\bibitem[{Chen et~al.(2017)Chen, Zhu, Ling, Wei, Jiang, and
  Inkpen}]{chen-etal-2017-enhanced}
Qian Chen, Xiaodan Zhu, Zhen-Hua Ling, Si~Wei, Hui Jiang, and Diana Inkpen.
  2017.
\newblock \href {https://doi.org/10.18653/v1/P17-1152} {Enhanced {LSTM} for
  natural language inference}.
\newblock In \emph{Proceedings of the 55th Annual Meeting of the Association
  for Computational Linguistics (Volume 1: Long Papers)}, pages 1657--1668,
  Vancouver, Canada. Association for Computational Linguistics.

\bibitem[{Chen et~al.(2020)Chen, Meng, Li, Chen, Xu, Xu, and
  Zhou}]{chen-etal-2020-bridging}
Xiuyi Chen, Fandong Meng, Peng Li, Feilong Chen, Shuang Xu, Bo~Xu, and Jie
  Zhou. 2020.
\newblock \href {https://doi.org/10.18653/v1/2020.emnlp-main.275} {Bridging the
  gap between prior and posterior knowledge selection for knowledge-grounded
  dialogue generation}.
\newblock In \emph{Proceedings of the 2020 Conference on Empirical Methods in
  Natural Language Processing (EMNLP)}, pages 3426--3437, Online. Association
  for Computational Linguistics.

\bibitem[{Cheng et~al.(2020)Cheng, Chang, Lee, and
  Toutanova}]{cheng-etal-2020-probabilistic}
Hao Cheng, Ming-Wei Chang, Kenton Lee, and Kristina Toutanova. 2020.
\newblock \href {https://doi.org/10.18653/v1/2020.acl-main.501} {Probabilistic
  assumptions matter: Improved models for distantly-supervised document-level
  question answering}.
\newblock In \emph{Proceedings of the 58th Annual Meeting of the Association
  for Computational Linguistics}, pages 5657--5667, Online. Association for
  Computational Linguistics.

\bibitem[{Cho et~al.(2014)Cho, Van~Merri{\"e}nboer, Gulcehre, Bahdanau,
  Bougares, Schwenk, and Bengio}]{cho2014learning}
Kyunghyun Cho, Bart Van~Merri{\"e}nboer, Caglar Gulcehre, Dzmitry Bahdanau,
  Fethi Bougares, Holger Schwenk, and Yoshua Bengio. 2014.
\newblock Learning phrase representations using rnn encoder-decoder for
  statistical machine translation.
\newblock \emph{arXiv preprint arXiv:1406.1078}.

\bibitem[{Dasigi et~al.(2019)Dasigi, Liu, Marasovi{\'c}, Smith, and
  Gardner}]{dasigi-etal-2019-quoref}
Pradeep Dasigi, Nelson~F. Liu, Ana Marasovi{\'c}, Noah~A. Smith, and Matt
  Gardner. 2019.
\newblock \href {https://doi.org/10.18653/v1/D19-1606} {{Q}uoref: A reading
  comprehension dataset with questions requiring coreferential reasoning}.
\newblock In \emph{Proceedings of the 2019 Conference on Empirical Methods in
  Natural Language Processing and the 9th International Joint Conference on
  Natural Language Processing (EMNLP-IJCNLP)}, pages 5925--5932, Hong Kong,
  China. Association for Computational Linguistics.

\bibitem[{Devlin et~al.(2019)Devlin, Chang, Lee, and
  Toutanova}]{devlin-etal-2019-bert}
Jacob Devlin, Ming-Wei Chang, Kenton Lee, and Kristina Toutanova. 2019.
\newblock \href {https://doi.org/10.18653/v1/N19-1423} {{BERT}: Pre-training of
  deep bidirectional transformers for language understanding}.
\newblock In \emph{Proceedings of the 2019 Conference of the North {A}merican
  Chapter of the Association for Computational Linguistics: Human Language
  Technologies, Volume 1 (Long and Short Papers)}, pages 4171--4186,
  Minneapolis, Minnesota. Association for Computational Linguistics.

\bibitem[{Dinan et~al.(2018)Dinan, Roller, Shuster, Fan, Auli, and
  Weston}]{dinan2018wizard}
Emily Dinan, Stephen Roller, Kurt Shuster, Angela Fan, Michael Auli, and Jason
  Weston. 2018.
\newblock Wizard of wikipedia: Knowledge-powered conversational agents.
\newblock In \emph{International Conference on Learning Representations}.

\bibitem[{Dinan et~al.(2019)Dinan, Roller, Shuster, Fan, Auli, and
  Weston}]{dinan2019wizard}
Emily Dinan, Stephen Roller, Kurt Shuster, Angela Fan, Michael Auli, and Jason
  Weston. 2019.
\newblock Wizard of wikipedia: Knowledge-powered conversational agents.
\newblock In \emph{International Conference on Learning Representations}.

\bibitem[{Feng et~al.(2021{\natexlab{a}})Feng, Patel, Wan, and
  Joshi}]{feng-etal-2021-multidoc2dial}
Song Feng, Siva~Sankalp Patel, Hui Wan, and Sachindra Joshi.
  2021{\natexlab{a}}.
\newblock \href {https://aclanthology.org/2021.emnlp-main.498}
  {{M}ulti{D}oc2{D}ial: Modeling dialogues grounded in multiple documents}.
\newblock In \emph{Proceedings of the 2021 Conference on Empirical Methods in
  Natural Language Processing}, pages 6162--6176, Online and Punta Cana,
  Dominican Republic. Association for Computational Linguistics.

\bibitem[{Feng et~al.(2021{\natexlab{b}})Feng, Reddy, Alikhani, He, Ji, Iyyer,
  and Yu}]{dialdoc-2021-document}
Song Feng, Siva Reddy, Malihe Alikhani, He~He, Yangfeng Ji, Mohit Iyyer, and
  Zhou Yu, editors. 2021{\natexlab{b}}.
\newblock \href {https://aclanthology.org/2021.dialdoc-1.0} {\emph{Proceedings
  of the 1st Workshop on Document-grounded Dialogue and Conversational Question
  Answering (DialDoc 2021)}}. Association for Computational Linguistics,
  Online.

\bibitem[{Ghazvininejad et~al.(2018)Ghazvininejad, Brockett, Chang, Dolan, Gao,
  Yih, and Galley}]{ghazvininejad2018knowledge}
Marjan Ghazvininejad, Chris Brockett, Ming-Wei Chang, Bill Dolan, Jianfeng Gao,
  Wen-tau Yih, and Michel Galley. 2018.
\newblock A knowledge-grounded neural conversation model.
\newblock In \emph{Proceedings of the AAAI Conference on Artificial
  Intelligence}, volume~32.

\bibitem[{He et~al.(2016)He, Zhang, Ren, and Sun}]{he2016deep}
Kaiming He, Xiangyu Zhang, Shaoqing Ren, and Jian Sun. 2016.
\newblock Deep residual learning for image recognition.
\newblock In \emph{Proceedings of the IEEE conference on computer vision and
  pattern recognition}, pages 770--778.

\bibitem[{Karpukhin et~al.(2020)Karpukhin, Oguz, Min, Lewis, Wu, Edunov, Chen,
  and Yih}]{karpukhin-etal-2020-dense}
Vladimir Karpukhin, Barlas Oguz, Sewon Min, Patrick Lewis, Ledell Wu, Sergey
  Edunov, Danqi Chen, and Wen-tau Yih. 2020.
\newblock \href {https://doi.org/10.18653/v1/2020.emnlp-main.550} {Dense
  passage retrieval for open-domain question answering}.
\newblock In \emph{Proceedings of the 2020 Conference on Empirical Methods in
  Natural Language Processing (EMNLP)}, pages 6769--6781, Online. Association
  for Computational Linguistics.

\bibitem[{Kim et~al.(2019)Kim, Ahn, and Kim}]{kim2019sequential}
Byeongchang Kim, Jaewoo Ahn, and Gunhee Kim. 2019.
\newblock Sequential latent knowledge selection for knowledge-grounded
  dialogue.
\newblock In \emph{International Conference on Learning Representations}.

\bibitem[{Kingma and Ba(2015)}]{kingma2015adam}
Diederik~P Kingma and Jimmy Ba. 2015.
\newblock Adam: A method for stochastic optimization.

\bibitem[{Lewis et~al.(2020{\natexlab{a}})Lewis, Liu, Goyal, Ghazvininejad,
  Mohamed, Levy, Stoyanov, and Zettlemoyer}]{lewis-etal-2020-bart}
Mike Lewis, Yinhan Liu, Naman Goyal, Marjan Ghazvininejad, Abdelrahman Mohamed,
  Omer Levy, Veselin Stoyanov, and Luke Zettlemoyer. 2020{\natexlab{a}}.
\newblock \href {https://doi.org/10.18653/v1/2020.acl-main.703} {{BART}:
  Denoising sequence-to-sequence pre-training for natural language generation,
  translation, and comprehension}.
\newblock In \emph{Proceedings of the 58th Annual Meeting of the Association
  for Computational Linguistics}, pages 7871--7880, Online. Association for
  Computational Linguistics.

\bibitem[{Lewis et~al.(2020{\natexlab{b}})Lewis, Perez, Piktus, Petroni,
  Karpukhin, Goyal, K{\"u}ttler, Lewis, Yih, Rockt{\"a}schel
  et~al.}]{lewis2020retrieval}
Patrick Lewis, Ethan Perez, Aleksandra Piktus, Fabio Petroni, Vladimir
  Karpukhin, Naman Goyal, Heinrich K{\"u}ttler, Mike Lewis, Wen-tau Yih, Tim
  Rockt{\"a}schel, et~al. 2020{\natexlab{b}}.
\newblock Retrieval-augmented generation for knowledge-intensive nlp tasks.
\newblock \emph{Advances in Neural Information Processing Systems},
  33:9459--9474.

\bibitem[{Li et~al.(2022)Li, Madhi~Namazifar, Bansal, Ji, Liu, and
  Hakkani-Tur}]{lienhanced}
Sha Li, Di~Jin Madhi~Namazifar, Mohit Bansal, Heng Ji, Yang Liu, and Dilek
  Hakkani-Tur. 2022.
\newblock Enhanced knowledge selection for grounded dialogues via document
  semantic graphs.

\bibitem[{Li and Liang(2021)}]{li-liang-2021-prefix}
Xiang~Lisa Li and Percy Liang. 2021.
\newblock \href {https://doi.org/10.18653/v1/2021.acl-long.353} {Prefix-tuning:
  Optimizing continuous prompts for generation}.
\newblock In \emph{Proceedings of the 59th Annual Meeting of the Association
  for Computational Linguistics and the 11th International Joint Conference on
  Natural Language Processing (Volume 1: Long Papers)}, pages 4582--4597,
  Online. Association for Computational Linguistics.

\bibitem[{Lian et~al.(2019)Lian, Xie, Wang, Peng, and Wu}]{lian2019learning}
Rongzhong Lian, Min Xie, Fan Wang, Jinhua Peng, and Hua Wu. 2019.
\newblock Learning to select knowledge for response generation in dialog
  systems.
\newblock In \emph{IJCAI International Joint Conference on Artificial
  Intelligence}, page 5081.

\bibitem[{Liu et~al.(2019)Liu, Niu, Wu, and Wang}]{liu-etal-2019-knowledge}
Zhibin Liu, Zheng-Yu Niu, Hua Wu, and Haifeng Wang. 2019.
\newblock \href {https://doi.org/10.18653/v1/D19-1187} {Knowledge aware
  conversation generation with explainable reasoning over augmented graphs}.
\newblock In \emph{Proceedings of the 2019 Conference on Empirical Methods in
  Natural Language Processing and the 9th International Joint Conference on
  Natural Language Processing (EMNLP-IJCNLP)}, pages 1782--1792, Hong Kong,
  China. Association for Computational Linguistics.

\bibitem[{Meng et~al.(2021)Meng, Ren, Chen, Ren, Xi, and
  Rijke}]{meng2021initiative}
Chuan Meng, Pengjie Ren, Zhumin Chen, Zhaochun Ren, Tengxiao Xi, and Maarten~de
  Rijke. 2021.
\newblock Initiative-aware self-supervised learning for knowledge-grounded
  conversations.
\newblock In \emph{Proceedings of the 44th International ACM SIGIR Conference
  on Research and Development in Information Retrieval}, pages 522--532.

\bibitem[{Meng et~al.(2020)Meng, Ren, Chen, Sun, Ren, Tu, and
  Rijke}]{meng2020dukenet}
Chuan Meng, Pengjie Ren, Zhumin Chen, Weiwei Sun, Zhaochun Ren, Zhaopeng Tu,
  and Maarten~de Rijke. 2020.
\newblock Dukenet: A dual knowledge interaction network for knowledge-grounded
  conversation.
\newblock In \emph{Proceedings of the 43rd International ACM SIGIR Conference
  on Research and Development in Information Retrieval}, pages 1151--1160.

\bibitem[{Moghe et~al.(2018)Moghe, Arora, Banerjee, and
  Khapra}]{moghe-etal-2018-towards}
Nikita Moghe, Siddhartha Arora, Suman Banerjee, and Mitesh~M. Khapra. 2018.
\newblock \href {https://doi.org/10.18653/v1/D18-1255} {Towards exploiting
  background knowledge for building conversation systems}.
\newblock In \emph{Proceedings of the 2018 Conference on Empirical Methods in
  Natural Language Processing}, pages 2322--2332, Brussels, Belgium.
  Association for Computational Linguistics.

\bibitem[{Moon et~al.(2019)Moon, Shah, Kumar, and
  Subba}]{moon-etal-2019-opendialkg}
Seungwhan Moon, Pararth Shah, Anuj Kumar, and Rajen Subba. 2019.
\newblock \href {https://doi.org/10.18653/v1/P19-1081} {{O}pen{D}ial{KG}:
  Explainable conversational reasoning with attention-based walks over
  knowledge graphs}.
\newblock In \emph{Proceedings of the 57th Annual Meeting of the Association
  for Computational Linguistics}, pages 845--854, Florence, Italy. Association
  for Computational Linguistics.

\bibitem[{Oono and Suzuki(2019)}]{oono2019graph}
Kenta Oono and Taiji Suzuki. 2019.
\newblock Graph neural networks exponentially lose expressive power for node
  classification.
\newblock In \emph{International Conference on Learning Representations}.

\bibitem[{Prabhumoye et~al.(2021)Prabhumoye, Hashimoto, Zhou, Black, and
  Salakhutdinov}]{prabhumoye-etal-2021-focused}
Shrimai Prabhumoye, Kazuma Hashimoto, Yingbo Zhou, Alan~W Black, and Ruslan
  Salakhutdinov. 2021.
\newblock \href {https://doi.org/10.18653/v1/2021.naacl-main.338} {Focused
  attention improves document-grounded generation}.
\newblock In \emph{Proceedings of the 2021 Conference of the North American
  Chapter of the Association for Computational Linguistics: Human Language
  Technologies}, pages 4274--4287, Online. Association for Computational
  Linguistics.

\bibitem[{Radford et~al.(2019)Radford, Wu, Child, Luan, Amodei, Sutskever
  et~al.}]{radford2019language}
Alec Radford, Jeffrey Wu, Rewon Child, David Luan, Dario Amodei, Ilya
  Sutskever, et~al. 2019.
\newblock Language models are unsupervised multitask learners.
\newblock \emph{OpenAI blog}, 1(8):9.

\bibitem[{Wolf et~al.(2020)Wolf, Debut, Sanh, Chaumond, Delangue, Moi, Cistac,
  Rault, Louf, Funtowicz, Davison, Shleifer, von Platen, Ma, Jernite, Plu, Xu,
  Le~Scao, Gugger, Drame, Lhoest, and Rush}]{wolf-etal-2020-transformers}
Thomas Wolf, Lysandre Debut, Victor Sanh, Julien Chaumond, Clement Delangue,
  Anthony Moi, Pierric Cistac, Tim Rault, Remi Louf, Morgan Funtowicz, Joe
  Davison, Sam Shleifer, Patrick von Platen, Clara Ma, Yacine Jernite, Julien
  Plu, Canwen Xu, Teven Le~Scao, Sylvain Gugger, Mariama Drame, Quentin Lhoest,
  and Alexander Rush. 2020.
\newblock \href {https://doi.org/10.18653/v1/2020.emnlp-demos.6} {Transformers:
  State-of-the-art natural language processing}.
\newblock In \emph{Proceedings of the 2020 Conference on Empirical Methods in
  Natural Language Processing: System Demonstrations}, pages 38--45, Online.
  Association for Computational Linguistics.

\bibitem[{Wu et~al.(2021)Wu, Lu, Hajishirzi, and
  Ostendorf}]{wu-etal-2021-dialki}
Zeqiu Wu, Bo-Ru Lu, Hannaneh Hajishirzi, and Mari Ostendorf. 2021.
\newblock \href {https://aclanthology.org/2021.emnlp-main.140} {{DIALKI}:
  Knowledge identification in conversational systems through dialogue-document
  contextualization}.
\newblock In \emph{Proceedings of the 2021 Conference on Empirical Methods in
  Natural Language Processing}, pages 1852--1863, Online and Punta Cana,
  Dominican Republic. Association for Computational Linguistics.

\bibitem[{Xu et~al.(2021{\natexlab{a}})Xu, Lei, Wang, Niu, Wu, and
  Che}]{xu2021enhancing}
Jun Xu, Zeyang Lei, Haifeng Wang, Zheng-Yu Niu, Hua Wu, and Wanxiang Che.
  2021{\natexlab{a}}.
\newblock Enhancing dialog coherence with event graph grounded content
  planning.
\newblock In \emph{Proceedings of the Twenty-Ninth International Conference on
  International Joint Conferences on Artificial Intelligence}, pages
  3941--3947.

\bibitem[{Xu et~al.(2021{\natexlab{b}})Xu, Lei, Wang, Niu, Wu, Che, Huang, and
  Liu}]{xu2021coherent}
Jun Xu, Zeyang Lei, Haifeng Wang, Zheng-Yu Niu, Hua Wu, Wanxiang Che, Jizhou
  Huang, and Ting Liu. 2021{\natexlab{b}}.
\newblock Coherent dialog generation with query graph.
\newblock \emph{Transactions on Asian and Low-Resource Language Information
  Processing}, 20(6):1--23.

\bibitem[{Xu et~al.(2020{\natexlab{a}})Xu, Wang, Niu, Wu, Che, and
  Liu}]{xu-etal-2020-conversational}
Jun Xu, Haifeng Wang, Zheng-Yu Niu, Hua Wu, Wanxiang Che, and Ting Liu.
  2020{\natexlab{a}}.
\newblock \href {https://doi.org/10.18653/v1/2020.acl-main.166} {Conversational
  graph grounded policy learning for open-domain conversation generation}.
\newblock In \emph{Proceedings of the 58th Annual Meeting of the Association
  for Computational Linguistics}, pages 1835--1845, Online. Association for
  Computational Linguistics.

\bibitem[{Xu et~al.(2020{\natexlab{b}})Xu, Wang, Niu, Wu, and
  Che}]{xu2020knowledge}
Jun Xu, Haifeng Wang, Zhengyu Niu, Hua Wu, and Wanxiang Che.
  2020{\natexlab{b}}.
\newblock Knowledge graph grounded goal planning for open-domain conversation
  generation.
\newblock In \emph{Proceedings of the AAAI Conference on Artificial
  Intelligence}, volume~34, pages 9338--9345.

\bibitem[{Xu et~al.(2019)Xu, Zhou, Gong, Liang, Tang, and Lin}]{xu2019end}
Lin Xu, Qixian Zhou, Ke~Gong, Xiaodan Liang, Jianheng Tang, and Liang Lin.
  2019.
\newblock End-to-end knowledge-routed relational dialogue system for automatic
  diagnosis.
\newblock In \emph{Proceedings of the AAAI Conference on Artificial
  Intelligence}, volume~33, pages 7346--7353.

\bibitem[{Ye et~al.(2020)Ye, Lin, Du, Liu, Li, Sun, and
  Liu}]{ye-etal-2020-coreferential}
Deming Ye, Yankai Lin, Jiaju Du, Zhenghao Liu, Peng Li, Maosong Sun, and
  Zhiyuan Liu. 2020.
\newblock \href {https://doi.org/10.18653/v1/2020.emnlp-main.582}
  {{C}oreferential {R}easoning {L}earning for {L}anguage {R}epresentation}.
\newblock In \emph{Proceedings of the 2020 Conference on Empirical Methods in
  Natural Language Processing (EMNLP)}, pages 7170--7186, Online. Association
  for Computational Linguistics.

\bibitem[{Zhan et~al.(2021{\natexlab{a}})Zhan, Shen, Chen, and
  Zhang}]{zhan-etal-2021-colv}
Haolan Zhan, Lei Shen, Hongshen Chen, and Hainan Zhang. 2021{\natexlab{a}}.
\newblock \href {https://aclanthology.org/2021.emnlp-main.172} {{C}o{LV}: A
  collaborative latent variable model for knowledge-grounded dialogue
  generation}.
\newblock In \emph{Proceedings of the 2021 Conference on Empirical Methods in
  Natural Language Processing}, pages 2250--2261, Online and Punta Cana,
  Dominican Republic. Association for Computational Linguistics.

\bibitem[{Zhan et~al.(2021{\natexlab{b}})Zhan, Zhang, Chen, Ding, Bao, and
  Lan}]{zhan-etal-2021-augmenting}
Haolan Zhan, Hainan Zhang, Hongshen Chen, Zhuoye Ding, Yongjun Bao, and Yanyan
  Lan. 2021{\natexlab{b}}.
\newblock \href {https://doi.org/10.18653/v1/2021.naacl-main.446} {Augmenting
  knowledge-grounded conversations with sequential knowledge transition}.
\newblock In \emph{Proceedings of the 2021 Conference of the North American
  Chapter of the Association for Computational Linguistics: Human Language
  Technologies}, pages 5621--5630, Online. Association for Computational
  Linguistics.

\bibitem[{Zhao et~al.(2020)Zhao, Wu, Xu, Tao, Zhao, and
  Yan}]{zhao-etal-2020-knowledge-grounded}
Xueliang Zhao, Wei Wu, Can Xu, Chongyang Tao, Dongyan Zhao, and Rui Yan. 2020.
\newblock \href {https://doi.org/10.18653/v1/2020.emnlp-main.272}
  {Knowledge-grounded dialogue generation with pre-trained language models}.
\newblock In \emph{Proceedings of the 2020 Conference on Empirical Methods in
  Natural Language Processing (EMNLP)}, pages 3377--3390, Online. Association
  for Computational Linguistics.

\bibitem[{Zheng et~al.(2020)Zheng, Cao, Jiang, and
  Huang}]{zheng-etal-2020-difference}
Chujie Zheng, Yunbo Cao, Daxin Jiang, and Minlie Huang. 2020.
\newblock \href {https://doi.org/10.18653/v1/2020.findings-emnlp.11}
  {Difference-aware knowledge selection for knowledge-grounded conversation
  generation}.
\newblock In \emph{Findings of the Association for Computational Linguistics:
  EMNLP 2020}, pages 115--125, Online. Association for Computational
  Linguistics.

\bibitem[{Zhou et~al.(2018{\natexlab{a}})Zhou, Young, Huang, Zhao, Xu, and
  Zhu}]{zhou2018commonsense}
Hao Zhou, Tom Young, Minlie Huang, Haizhou Zhao, Jingfang Xu, and Xiaoyan Zhu.
  2018{\natexlab{a}}.
\newblock Commonsense knowledge aware conversation generation with graph
  attention.
\newblock In \emph{IJCAI}, pages 4623--4629.

\bibitem[{Zhou et~al.(2018{\natexlab{b}})Zhou, Prabhumoye, and
  Black}]{zhou-etal-2018-dataset}
Kangyan Zhou, Shrimai Prabhumoye, and Alan~W Black. 2018{\natexlab{b}}.
\newblock \href {https://doi.org/10.18653/v1/D18-1076} {A dataset for document
  grounded conversations}.
\newblock In \emph{Proceedings of the 2018 Conference on Empirical Methods in
  Natural Language Processing}, pages 708--713, Brussels, Belgium. Association
  for Computational Linguistics.

\end{thebibliography}

\appendix
\label{sec:appendix}

\section{Implementation Details}
We set the maximum lengths of model input to 512, which is also the longest input limit for the BERT model, in order to fit the longer passage text as much as possible on both datasets. We employ a Linear layer to transform the output features of BERT from 768 to 320 to reduce memory usage. The edge embedding size is set to 64. The hidden size and headers of RES-RGAT are 1024 and 8 respectively while the alpha value of Graph Attention Network is 0.2. We utilize a unidirectional stacked GRU model for Differential Sequential Learning, the number of GRU layers is 2. 

For response generation, we apply PrefixTunning~\cite{li-liang-2021-prefix} on BART~\cite{lewis-etal-2020-bart} large model to learn the responses generation model based on the knowledge selection results from the previous stage. We use the prefix length 200 and the hidden dimension of 800 for all the methods using PrefixTuning generator. The PrefixTuning generator takes about 4 hours and 30 epoch to become converged during training on 4 V100 32G GPUs, which is much faster and more resource saving than fine-tuning BART large.

\section{Dataset Processing Details}
\noindent\textbf{WoW. }
There are more than 130k different documents from Wikipedia in WoW training set. We keep 350 high-frequency relations from the Wiki knowledge graph, covering these 130k documents. The top-10 wiki relations with corresponding frequency are shown as follows:
\begin{enumerate}
\item ('subclass of', 27015)
\item ('facet of', 11381)
\item ('sport', 10646)
\item ('performer', 9482)
\item ('part of', 6892)
\item ('manufacturer', 5742)
\item ('instance of', 5551)
\item ('history of topic', 5517)
\item ('has part', 5445)
\item ('follows', 5077)
\end{enumerate}

As shown in Table~\ref{tab:average-relations}, for topic relations, we found the \texttt{word\_overlap} edges is denser than the commonsense edges from wikiData, giving the average edge number of 8.11 and 2.89, respectively. While for knowledge relations, the \texttt{coreference\_link} has much less average number of relations in one sample than other two types relations, which again proves that \texttt{coreference\_link} with more accurate knowledge relations lead to better knowledge selection results without introducing wrong structures information to CorefDiffs framework.

\begin{table}[hbt]
\centering
\resizebox{0.99\linewidth}{!}{
\begin{tabular}{lcc|ccc}
\toprule
\multirow{2}{*}{} & \multicolumn{2}{c}{\textbf{Topic Relations}}& \multicolumn{3}{c}{\textbf{Knowledge relations}}\\ 
                  & \textit{WordOverlap} & \textit{WikiGraph} & \textit{Partial} & \textit{EntityLink} & \textit{Coreference} \\ 
\midrule
Freq         &    8.11     &  2.89   &  61.18  &    87.52       &   15.90 \\
\bottomrule
\end{tabular}
}
\caption{Average number of different kinds of relations in one sample on the WoW training set.}
\label{tab:average-relations}
\end{table}

\noindent\textbf{Holl-E. }
Different from WoW, each sample of Holl-E has only one topic, which is the movie in this session of conversation. There are four types of information for each movie in Holl-E, which are plots, comments, reviews, and table information. So we simply divide all the knowledge sentences of each movie into four topics. As the absence of common sense relations of such topics in Holl-E, we count the co-occurrence relationship of all topics in the training set as the relations between topics in Holl-E. The relations between knowledge are as same as the WoW, using coreference relations in passage text. The relations between knowledge and topics are sentence order of knowledge sentence in the original text, which is also used in WoW. 

\noindent\textbf{CMU-DoG. }
CMU-DoG is a document-grounded conversation dataset about movie, which is the same as Holl-E. The difference is that CMU-DoG includes only one grounding document(one topic) at each dialog turn. The relations of topics is absent as there is only one topic in grounding document. The relations between knowledge are as same as the WoW and Holl-E with coreference relations. The relations between knowledge and topics are sentence order of knowledge sentence in the passage, which is also consistent with WoW and Holl-E. On the other hand, CMU-DoG didn't contain the gold knowledge of knowledge selection task. We adopt unigram F1 score as similarity function, selecting the knowledge closest to the ground-truth response as gold knowledge to train the knowledge selection model.

\noindent\textbf{MultiDoc2Dial. }
MultiDoc2Dial includes multiple grounding documents at each dialog turn. We construct the graph following the steps of Holl-E. However, MultiDoc2Dial introduces a span prediction task to locate knowledge set in the original document instead of knowledge selection. But that's ok, it easy for our framework to transfer downstream task by using two independent classifier to predict both start knowledge segment and end knowledge segment instead of one classifier for knowledge selection. Simultaneously, we replace the metric from knowledge accuracy to knowledge EM, which is used in MultiDoc2Dial. For convenience, we still use \textbf{KL} in Table~\ref{tab:cmudog_multidoc2dial_results} to denote the EM metric in MultiDoc2Dial.

\section{Analysis on Partial Order Edge}
\label{apd:partial_order}
\begin{figure}[h]
\centering
  \includegraphics[width=0.7\linewidth]{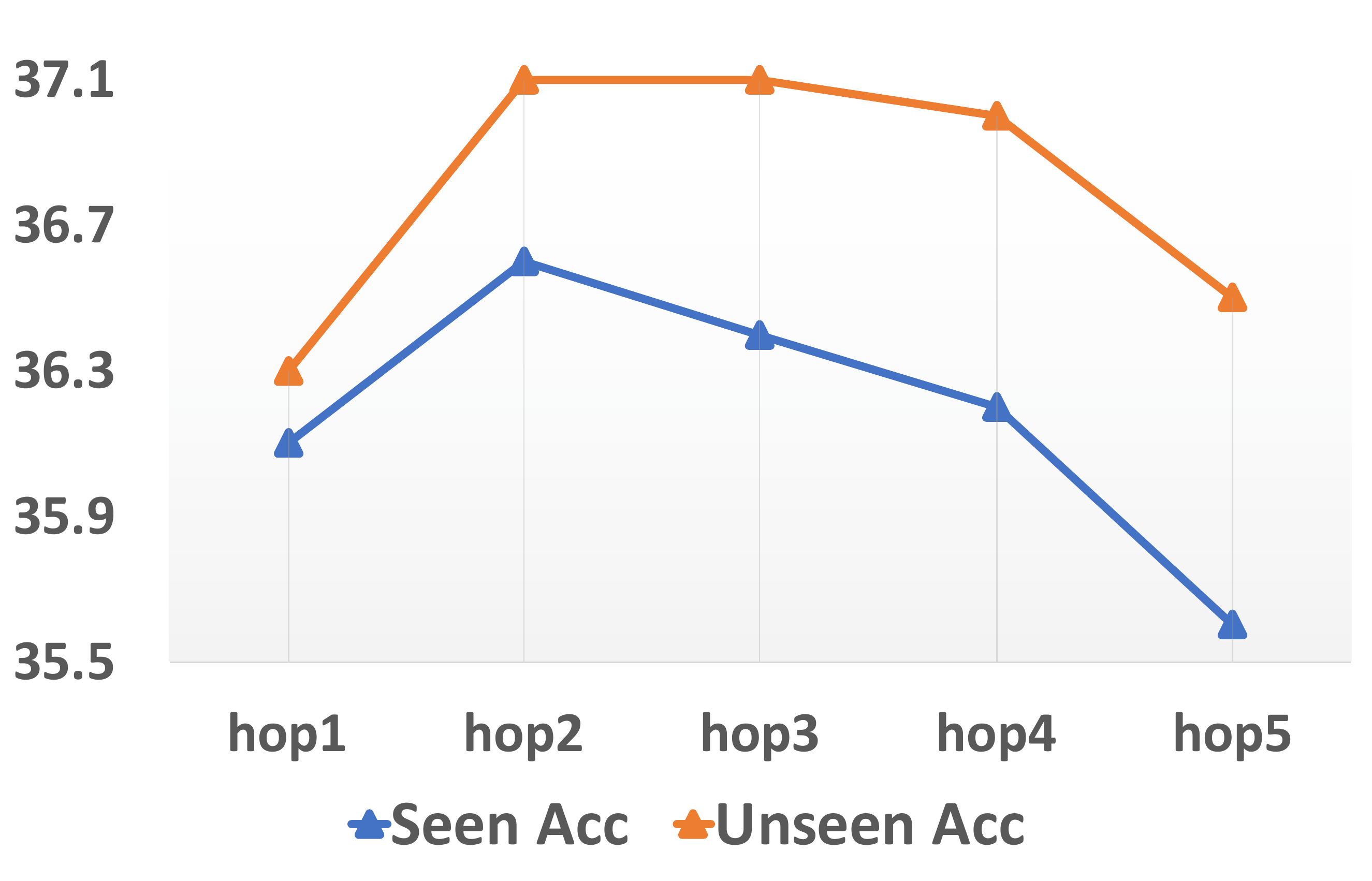}
\caption{Knowledge accuracy for partial order with different hops.}

\label{fig:partial_hop}
\end{figure}

For partial order relations, we explored the effects of different hops. Hop-k partial order relation means each knowledge vertex is connected with $k$ knowledge vertices behind according to the sentence order. As shown in Fig~\ref{fig:partial_hop}, hop-2 partial relation performed the best. A hop that was too large or too small could cause information loss or introduce many erroneous connections.

\end{document}